\documentclass[10pt,twocolumn,letterpaper]{article}

\usepackage{iccv}
\usepackage{times}
\usepackage{epsfig}
\usepackage{graphicx}
\usepackage{amsmath}
\usepackage{amssymb}
\usepackage{booktabs}
\usepackage{dblfloatfix}
\DeclareMathOperator*{\detect}{Detect}
\usepackage[pagebackref=true,breaklinks=true,letterpaper=true,colorlinks,bookmarks=false]{hyperref}
\usepackage{authblk}

\makeatletter
\renewcommand\AB@affilsepx{ \protect\Affilfont}
\makeatother
\iccvfinalcopy 

\ificcvfinal\fi
\begin{document}

\title{OMNIA Faster R-CNN: \\ Detection in the wild through dataset merging and soft distillation}

\author[1]{Alexandre Rame
\thanks{\tt\small alexandre.rame@heuritech.com}
}
\author[1]{Emilien Garreau
\thanks{\tt\small emilien.garreau@heuritech.com}
}
\author[1,2]{Hedi Ben-Younes
\thanks{\tt\small hedi.benyounes@gmail.com}
}
\author[1]{Charles Ollion
\thanks{\tt\small ollion@heuritech.com}
}
\affil[1]{Heuritech}
\affil[2]{Sorbonne University, CNRS, LIP6}

\maketitle

\begin{abstract}

Object detectors tend to perform poorly in new or open domains, and require exhaustive yet costly annotations from fully labeled datasets. We aim at benefiting from several datasets with different categories but without additional labelling, not only to increase the number of categories detected, but also to take advantage from transfer learning and to enhance domain independence.

Our dataset merging procedure starts with training several initial Faster R-CNN on the different datasets while considering the complementary datasets' images for domain adaptation. Similarly to self-training methods, the predictions of these initial detectors mitigate the missing annotations on the complementary datasets. The final \textbf{OMNIA Faster R-CNN} is trained with all categories on the union of the datasets enriched by predictions. The joint training handles unsafe targets with a new classification loss called \textbf{SoftSig} in a softly supervised way.

Experimental results show that in the case of fashion detection for images in the wild, merging \textit{Modanet} with \textit{COCO} increases the final performance from 45.5\% to 57.4\% in mAP. Applying our soft distillation to the task of detection with domain shift between \textit{GTA} and \textit{Cityscapes} enables to beat the state-of-the-art by 5.3 points. Our methodology could unlock object detection for real-world applications without immense datasets.

\end{abstract}

\section{Introduction}

\paragraph{}

Convolutional Neural Networks (CNNs) \cite{krizhevsky2012imagenet, he2016deep} has become the default method for any computer vision task, and is widely used for problems such as image classification, semantic segmentation or visual relationship detection. One of the key computer vision task is certainly object detection. This task aims at localizing specific objects in an image. Best-performing detectors are Fully Supervised Detectors (FSDs): instance annotations are needed for each object in each image, composed of a category and its location.

\begin{figure}[t!]
\centering
\includegraphics[width=0.5\textwidth]{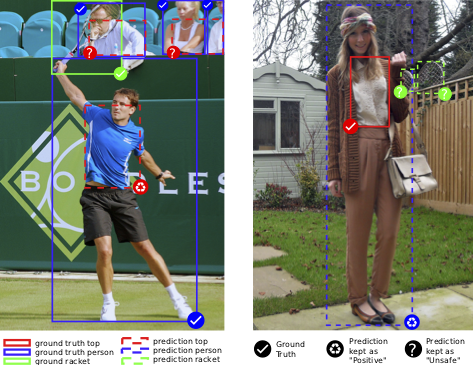}
\label{fig:cocomodanet_schema}
\caption{Dataset merging between \textit{COCO} (left) and \textit{Modanet} (right). We fill missing annotations with predictions.}
\end{figure}

As it has been remarked in \cite{wu2018soft, rosenberg2005semi, mnih2012learning, reed2014training}, FSDs are sensitive to noisy and missing annotations. In particular, the performance of these models are deteriorated when categories are not labeled in some images. Thus, the cost of adding one new category in a dataset is very high: a manual enrichment throughout the whole dataset is required to find all the occurrences of this category. Moreover, the instance annotation process for object detection is expensive and time-consuming. For these reasons, most detection datasets are constrained to a small set of object categories, and are limited in size compared to datasets for other tasks.

The problem of training CNNs for classification on small datasets is usually tackled by transfer learning methods \cite{sharif2014cnn}: the network is pretrained on a large labeled dataset such as \textit{Imagenet} \cite{deng2009imagenet}, and fine-tuned on the task at hand while preserving the original capabilities \cite{li2017learning}. On the detection task, transfer learning methods have also been investigated using \textit{COCO} \cite{lin2014microsoft} for pretraining in Mask R-CNN \cite{he2017mask}. However, these methods are only efficient when the classes in the smaller dataset are labeled in the larger. There are no clear approaches on how to best use \textit{COCO} to detect new unknown categories, such as fashion garments.

For example, the fashion dataset \textit{Modanet} \cite{zheng2018modanet} seems adequate to learn how to detect garments on social media. However social media images are unconstrained whereas \textit{Modanet} images are fashion oriented. The distribution mismatch and the considerable domain shift between the two domains lead to a significant performance drop \cite{beery2018recognition}, and domain adaptation (DA) is needed \cite{raj2015subspace, chen2018domain, inoue2018cross, shan2018pixel}. But we would also like to benefit from the labeled data in \textit{COCO} \cite{lin2014microsoft}. Being able to detect \textit{bottles} or \textit{fire hydrants} improves detection scores on \textit{dresses} by reducing false positive predictions (see Figure  \ref{fig:modanetvsomnia}). However, categories in \textit{COCO} and \textit{Modanet} are not the same (only \textit{bag} and \textit{tie} are in both). If a Faster R-CNN is trained on the naive concatenation of the two datasets, \textit{dresses} instances in \textit{COCO}'s images are considered as background and deteriorate the performances. Multitasking is straightforward in classification \cite{caruana1997multitask, chapelle2011boosted} but as far as we know currently impossible for detectors without full relabelling.

\paragraph{}

In this article, we take a step back from classical transfer learning approaches. We instead tackle the problem from the angle of \textbf{dataset merging}. We propose a framework to train detectors that are robust on all categories from all merged datasets -- even though each dataset is annotated on distinct sets of categories. We call this framework \textbf{OMNIA Faster R-CNN}. By widening the scope of the detector, our training procedure provides a better understanding of the background, and increases the detector's accuracy. We are \textbf{detecting without forgetting} because we train on several datasets simultaneously. All available labeled data are used in our proposed procedure with no additional labelling. 

Our distinctive approach is to \textbf{alleviate missing annotations with model predictions as weak supervision} (see Figure \ref{fig:cocomodanet_schema}). To avoid a new labelling step, we will instead use predicted \textit{dresses} (with our FSD trained on \textit{Modanet}) as targets on \textit{COCO}. The final detector needs to take into account the notion of uncertainty of our predictions, by considering ground truths and safe predictions (with high confidence) differently from unsafe predictions (with low confidence).

The contributions of this paper are threefold: (1) we introduce a dataset merging procedure that enables the fusion of several datasets with heterogeneous categories: without any additional labelling, we benefit from large open source detection datasets (2) our self training approach handles \textbf{unsafe predictions} with a custom classification loss \textbf{SoftSig} (3) we combine self-training and data distillation for domain adaptation.

We prove than our simple procedure has high practical impact for numerous applications such as fashion and autonomous driving. Merging \textit{COCO} \cite{lin2014microsoft} and \textit{Modanet} \cite{zheng2018modanet} increases performance for object detection of fashion garments on \textit{OpenImages} \cite{openimages} from 45.5\% to 57.4\% in mAP@0.5. For ablation study, we use the synthetic dataset \textit{SIM10k} to learn how to detect \textit{cars} on real-world images, without using the \textit{car} annotations from \textit{Cityscapes}. Appending a new category to a detection dataset is a special case of the more general problem of dataset merging. Performances are close to those obtained with full annotations. Finally, our soft distillation procedure can be applied to detection with domain adaptation, and beat the state-of-the-art by 5.3 points, from 39.6 to 44.9 on the task of learning from \textit{SIM10k} to \textit{Cityscapes}.

\section{Related work}

\paragraph{Object detection}

Many approaches have been developed recently for fast and accurate detection \cite{liu2016ssd, redmon2017yolo9000, dai2016r, lin2017focal, law2018cornernet, girshick2014rich, girshick2015fast, ren2015faster,he2017mask}. We selected the \textbf{Faster R-CNN} framework \cite{ren2015faster} for its state-of-the-art performance, and for better comparability since this is the architecture mostly used in the context of domain adaptation detection. It introduces the RPN, which aims at generating class-agnostic object proposals by classifying whether the Region of Interest (ROI) is a background or a foreground. The final network then classifies the proposed ROIs into object categories or background, while refining the coordinates. The Faster R-CNN is trained end-to-end by multitasking. An important feature is the fixed background class that encapsulates real background classes (such as \textit{sky}, \textit{sun} etc) and all unlabeled other categories. It nonetheless has some failures, \eg the classification depends a lot on the context and can easily be fooled by contextual overfitting \cite{cheng2018revisiting, rosenfeld2018elephant}.

Learning visual representations requires \textbf{large scale training data} to provide good coverage over the space of possible appearances. Classification datasets \cite{deng2009imagenet, torralba200880} with millions of images and numerous categories enabled the boost in accuracies of CNNs \cite{sun2017revisiting}. However, the biggest fully supervised detection and segmentation datasets \cite{lin2014microsoft, zheng2018modanet, johnson2016driving, cordts2016cityscapes} are still on the order of hundreds of thousands of images. That's why many detection approaches tried to benefit from other type of datasets. Huge classification datasets are used for pretraining \cite{sun2017revisiting}, or even during training to expand the number of classes detected in Yolo9000 \cite{redmon2017yolo9000}, by using image-level class annotations as weak supervision for object detection. Image-level labels are used to reduce the labelling time, but lead to many missing instances in \textit{OpenImages} \cite{openimages}. Some recent approaches leverage semantic embedding from text to detect object classes for Zero-Shot Object Detection \cite{bansal2018zero, rahman2018zero, zhu2019zero}. As far as we know, Kemnitz \textit{et al.} \cite{kemnitz2018combining} is the only approach for merging detection datasets with heterogeneous label subsets (for semantic segmentation of medical images).

Computer vision setup usually assumes that the categories are the same in training and testing. The notion of \textit{third class} was first introduced in Universum \cite{weston2006inference} by Weston \textit{et al.}, and is really similar to our background class. The practical approach to better understand the background is to collect samples that hopefully represent the unlabeled categories. However, it is very difficult to sample enough background regions that represent all \textit{other} categories. In \textbf{novelty detection}, recent approaches \cite{busto2017open, saito2018open, kliger2018novelty} tried to detect these new categories automatically. In detection, \cite{bansal2018zero} aims at defining a large open vocabulary for differentiating various background regions in order to spread the background classes over the embedding space. The problem of detecting a variable and large number of categories in open set images has not received much attention but requires strong generalization skills.

\paragraph{Self-training}

Our paper is related to self-training, a strategy in which the predictions of a model on unlabeled data are used to train itself iteratively \cite{yarowsky1995unsupervised, nigam2000analyzing, chen2013neil, li2018mathcal,blum1998combining}. It relies on accurate predictions of the initial model to generate correct new training examples. It was applied to object detection \cite{rosenberg2005semi, radosavovic2017data, akram2018leveraging}: large unlabeled datasets improved performance on object and human keypoint detection tasks. Model distillation \cite{hinton2015distilling, lopez2015unifying, chen2017learning, li2018mathcal} aggregates the inferences from multiple models by ensembling \cite{hansen1990neural}. Data distillation by Radosavovic \textit{et al.} \cite{radosavovic2017data} aggregates the inferences of multiple transformations of a data point for keypoint detection. These two previous distillations can be combined  \cite{min2018robust}. To decide which predictions should be integrated as ground truths, Papadopoulos \textit{et al.} \cite{papadopoulos2016we} added an human-in-the-loop while Rosenberg \textit{et al.} \cite{rosenberg2005semi} used an external selection metric. In most of them, unlabeled data are not sampled from a different distribution. Only Dai \textit{et al.} \cite{dai2018dark} proposed a solution highly specific to their datasets: the Gradual Model Adaptation that progressively integrates images further from the source domain, building a bridge to their target domain.

In this self-training procedure, some instances will be missing and may cause \textbf{omission noise} on the final classifier. Mnih \textit{et al.} \cite{mnih2012learning} propose the asymmetric Bernoulli noise model. Reed \textit{et al.} \cite{reed2014training} use a consistency objective to reduce the loss. Finally, Radosavovic \textit{et al.} \cite{radosavovic2017data} claim that it does not affect their performance. \cite{wu2018soft} noticed a small but noticeable gap of performances when annotations are missing, and introduced a soft sampling so that unsafe background regions are given a smaller weight compared to the hard negatives background. They also highlighted that sampling true hard negatives is essential for robust learning. Our experiments confirmed these 2 analysis: bridging this gap while sampling hard negatives is our main challenge.

\paragraph{Domain adaptation}

Recognition algorithms assume that training and testing data are drawn from similar distributions, but it is often false in practice. The domain shift can be caused by differences in resolutions, view points, backgrounds, illuminations or even image qualities. Numerous \textbf{classification} approaches were surveyed in \cite{wang2018deep}. Domain adaptation can be optimized in an adversarial training manner with a Gradient Reversal Layer (GRL). Ganin and Lempitsky \cite{ganin2014unsupervised} introduced this layer to force the CNN to maximize a domain classification loss while minimizing the usual label prediction loss \cite{long2015learning, saito2018open, tzeng2017adversarial}.

Domain adaptation has been studied only recently for problems such as \textbf{detection} \cite{tang2012shifting, xu2014domain, hattori2015learning, raj2015subspace, chen2018road, zhang2017curriculum}. Inoue \textit{et al.} \cite{inoue2018cross} proposed a progressive domain adaptation technique by fine-tuning on two types of artificially and automatically generated samples: their initially unlabeled data come from another domain but have image-level labels. Finally, in this paper we mainly use Domain Adaptive Faster R-CNN built by Chen \textit{et al.} \cite{chen2018domain}, that addresses the semantic domain shift. They added 3 new terms in the training loss: (1) an instance level domain adaptation loss, to align the ROIs features distribution (2) an image level domain adaptation loss, to eliminate the domain distribution mismatch on the image level (3) a consistency regularization. The adversarial training is achieved with a GRL \cite{ganin2014unsupervised}. Recently, Shan \textit{et al.} \cite{shan2018pixel} additionally challenge the pixel domain shift by employing consistency losses from CycleGAN \cite{zhu2017unpaired}. They were limited in number of possible domains but were the state-of-the-art up to now. 
\section{Approach}

\paragraph{}

\begin{figure}[!t!]
\centering
\includegraphics[width=0.5\textwidth]{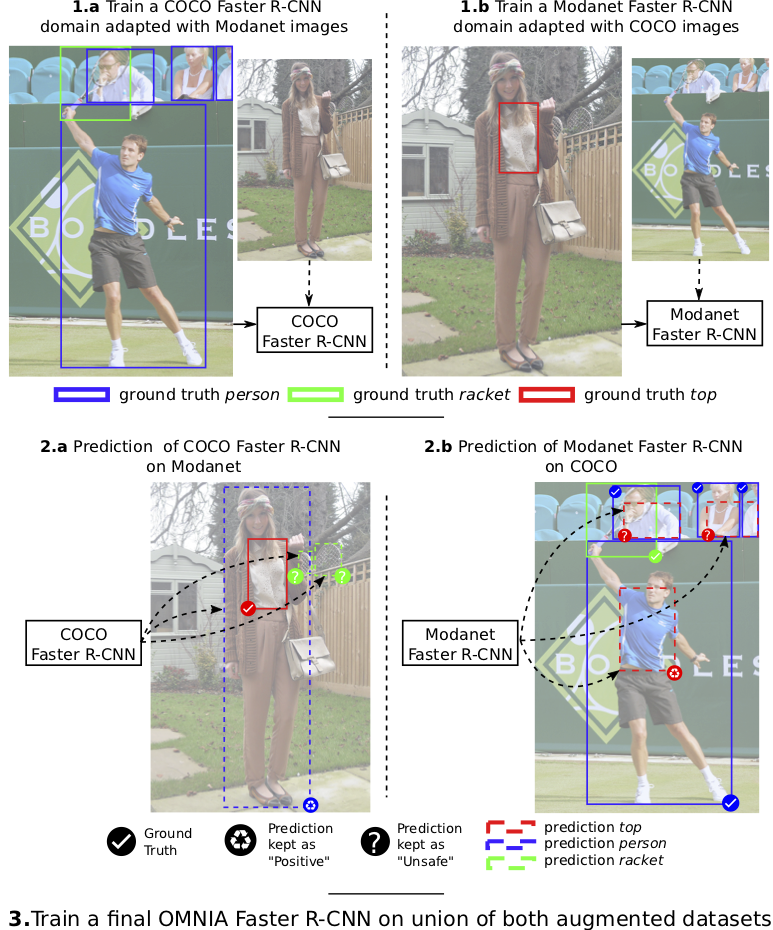}
\caption{Dataset merging procedure, leveraging Faster R-CNN, domain adaptation and self-training. Best seen in color.}
\label{fig:proceduremerging}
\end{figure}

Our procedure is illustrated in Figure \ref{fig:proceduremerging}. We combine several labeled datasets to build one common detector that aggregates all available information and detects everything: the \textbf{OMNIA Faster R-CNN}.

\subsection{Initial Trainings}

Two datasets, $\mathcal{D}_a$ (defined by categories $\mathcal{C}_a$, images $\mathcal{I}_a$ and ground truths $\mathcal{G}_a$), and $\mathcal{D}_b$ (defined by categories $\mathcal{C}_b$, images $\mathcal{I}_b$ and ground truths $G_b$) are provided.
As each dataset has its own bias, we train two different DA Faster R-CNN \cite{chen2018domain} domain adapted to the complementary unlabeled images. The first DA Faster R-CNN $\detect_a$ learns to detect $\mathcal{C}_a$ on $\mathcal{I}_a$, and has access to images $\mathcal{I}_b$ (without seeing $\mathcal{G}_b$) to minimize the semantic shift between the  features extracted from the two domains. Similarly, the second DA Faster R-CNN $\detect_b$ learns to detect $\mathcal{C}_b$ on $\mathcal{I}_b$, and has access to images $\mathcal{I}_a$ (without seeing $\mathcal{G}_a$). This adaptation aims at making $\detect_a$ (respectively $\detect_b$) as accurate as possible on $\mathcal{I}_b$ (respectively on $\mathcal{I}_a$).

\subsection{Merging Procedure}

\paragraph{Self-training}

We aim at training a detector simultaneously for all categories $\mathcal{C}_{a} \cup \mathcal{C}_b$ on all images $\mathcal{I}_{a} \cup \mathcal{I}_{b}$. $\mathcal{G}_{a}$ provides instances annotations for images from $\mathcal{I}_a$, but only for categories from $\mathcal{C}_{a}$. Ideally we would like to have access to the ground truths of categories from $\mathcal{C}_b \setminus \mathcal{C}_a$ for images in $\mathcal{I}_a$. The predictions of $\detect_b$ on $I_a$ will replace a new expensive labelling step. For the next training on categories from $\mathcal{C}_{a} \cup \mathcal{C}_b$, the new targets for $I_a$ are $\mathcal{G}_a \cup \detect_{b}(\mathcal{I}_{a})$.

\paragraph{Prediction Selection}
We want to alleviate the fact that some predictions in $\detect_{b}(\mathcal{I}_{a})$ are erroneous. A prediction with a high detection score is more trustworthy than another one with a lower score: the classification predicted score can be used as proxy for annotation quality \cite{radosavovic2017data}. We only take into account predictions with a score higher than $threshold\_low$. Another threshold $threshold\_high$ is defined: all predictions with a higher score will be considered as ground truths.
\begin{itemize}
  \item $ score \leq threshold\_low$, the prediction is discarded
  \item if $score > threshold\_high$, the instance is a safe prediction and will be considered as a ground truth
  \item $threshold\_low < score \leq threshold\_high$, the instance is an unsafe prediction
\end{itemize}
Furthermore, if a prediction has a very high IoU overlap with a human labeled ground truth, we assume that the initial detector made a mistake and we simply discard the prediction.
The same procedure is applied to images from $I_b$.

\subsection{OMNIA Faster R-CNN}

We introduce our \textbf{OMNIA Faster R-CNN}, trained on union on both augmented datasets and that handles unsafe predictions: some of them are correct and should not be considered as background.

\subsubsection{RPN}

To train the RPN, a binary class label for foreground/background classification is usually assigned to each anchor. In our custom RPN, we add a new \textit{undefined} label. There are now 3 possibilities:
\begin{itemize}
    \item a \textit{positive} label is given if the anchor matches with \textbf{any} ground truth or safe prediction, i.e it has an IoU overlap higher than a certain threshold
    \item an \textit{undefined} label is given if the anchor matches with an unsafe prediction
    \item a \textit{negative} label is given if the anchor has a max IoU overlap lower than a certain threshold with \textbf{all} ground truths and safe predictions
\end{itemize}
\textit{Positive} and \textit{negative} anchors are sampled given a fixed probability. In conclusion, the anchors that match with unsafe predictions will contribute neither to the classification loss nor to the regression loss of the RPN.

\subsubsection{SoftSig Box Classifier}

\paragraph{}

The Box classification loss needs to handle the ROIs that match with unsafe predictions. First, they will not contribute to the regression loss. In the classification loss, considering them as background regions would be too conservative in terms of exploration and may be equivalent to having lot of missing annotations. On contrary, discarding them totally from the loss would not exploit all the available information. 
In particular, even though we are not fully confident that the category $c$ is true for this ROI, it is very likely that all other classes are false. Indeed, the network should only predict either background or $c$, but not any other category.
We propose a simple and efficient custom classification function that takes advantage of all the available information during the training. We named this loss \textbf{SoftSig} because it combines \textbf{Soft}max and \textbf{Sig}moid activation fonctions while handling \textbf{Soft Sig}ns.
It is a mixed loss between a masked categorical and binary cross-entropy.

\paragraph{Notations}
We consider that we have $C$ categories (plus the background) and that we sample $R$ ROIs. $\forall 0 \leq r < R, 0 \leq c \leq C$, the target $t_{r}^{c} = 1$ iif $r$ matches with a box of category $c$. $x_{r}^{c}$ is the logit (the network prediction before any activation) at category $c$ for ROI $r$.

\paragraph{Masked Categorical Cross-Entropy}
The categorical cross-entropy aligns the predicted probabilities with the targets. It is the standard loss for classification, where all categories compete again each others. ROIs that match an unsafe prediction are completely discarded, and their associated loss term is multiplied by a binary variable $m_{r}$: if $r$ matches an unsafe prediction, $m_r=0$, else $m_r=1$. 

\begin{equation}
\label{"eq:categoricalcrossentropy"}
L_{categorical} = - \frac{1}{R} \sum_{r} m_{r} \sum_{c} t_{r}^{c} \log (\mathit{softmax}(x_{r}^{c}))
\end{equation}

\paragraph{Masked Binary Cross-Entropy}
Binary cross-entropy treats the different categories as independent binary classification tasks, where each $x_{r}^{c}$ decides whether the sample belongs to a category $c$ independently from the other categories. We slightly modify its formula to take into account the ROIs that match with unsafe predictions. The terms of the binary cross-entropy are weighted by a category-dependent binary mask $w_r^c \in \{0, 1 \}$. If the ROI matches an unsafe prediction of category $c$, we set $w_r^c = w_r^{background} = 0$. In all other cases, we set $w_r^c=1$.

\begin{equation}
\begin{split}
L_{binary} =& - \frac{1}{R(C + 1)} \sum_{r, c} w_r^c (t_r^c \log (\mathit{sigmoid}(x_r^c)) + \\
&(1-t_r^c) \log (1-\mathit{sigmoid}(x_r^c)))
\end{split}
\label{eq:binaryentropy}
\end{equation}

If ROI $r$ matches with an unsafe prediction of category $c$, $x_{r}^{c}$ and $x_{r}^{background}$ will not contribute to the loss from Equation (\ref{eq:binaryentropy}) but the logits for other categories will be decreased after an optimization step. The detector will only propagate safe gradients. The network is trained to predict either background or $c$.

\paragraph{Softsig} Finally we use the sum of  (\ref{"eq:categoricalcrossentropy"}) and (\ref{eq:binaryentropy}) as our final classification loss
\begin{equation}
  \label{eq:fastR-CNN_loss}
  L_{softsig} = L_{categorical} + \lambda_{binary} L_{binary}
\end{equation}
where $\lambda_{binary}$ is a trade-off parameter to balance between the two individual losses so that they contribute similarly.

\section{Experiments}

\begin{figure*}[h]
    \centering
    \includegraphics[width=\textwidth]{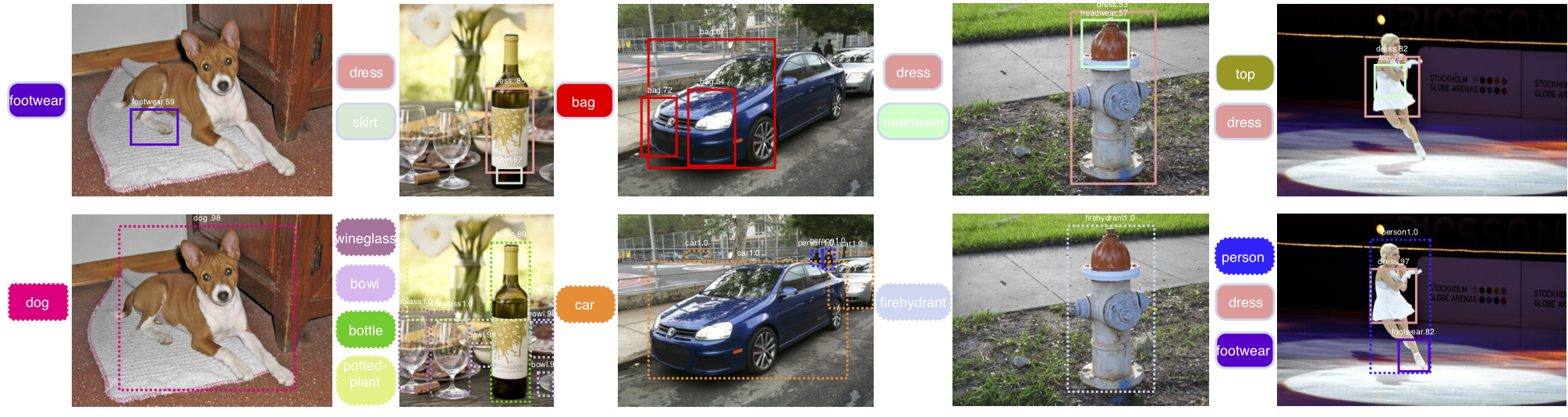}
    \caption{Predictions comparison between \textit{Modanet} Faster R-CNN (top) and \textbf{OMNIA Faster R-CNN} (bottom) on \textit{OpenImages}. Our procedure reduces the number of hard false positives for fashion objects. Dotted boxes represent \textit{COCO} categories whereas the solid lines are for \textit{Modanet} garments. Best seen in color.}
    \label{fig:modanetvsomnia}
\end{figure*}

\begin{table*}

\resizebox{2.2\columnwidth}{!}{%
  \begin{tabular}{| c | c c c c c | c c || c c | c c c |}

\hline
\multicolumn{1}{|c||}{} & \multicolumn{7}{|c||}{Contributions} & \multicolumn{5}{c||}{Results}\\
\hline
\multicolumn{1}{|c||}{} & \multicolumn{5}{|c||}{Dataset Creation Procedure} & \multicolumn{2}{|c||}{Handling Unsafe Predictions} & \multicolumn{2}{c||}{Mean} & \multicolumn{3}{c||}{Per Category} \\
\hline
Method & DA & Merging & Self training & Source & Target & Masking & Binary & \begin{tabular}{@{}c@{}}mean \\ mAP/MoLRP\end{tabular} & \begin{tabular}{@{}c@{}}mean w/o bag\&tie \\ mAP/MoLRP\end{tabular} & 
\begin{tabular}{@{}c@{}}dress \\ AP/oLRP\end{tabular} &
\begin{tabular}{@{}c@{}}footwear \\ AP/oLRP\end{tabular} &
\begin{tabular}{@{}c@{}}pants \\ AP/oLRP\end{tabular} \\
\hline
\hline
Domain Adaptive \cite{chen2018domain} & \checkmark & & & $S_{source}$ $U_{target}$ & $U_{source}$ $U_{target}$ & & & 45.5/77.8 & 48.9/75.6 & 67.5/66.8 & 19.8/91.9 & 42.6/78.1 \\\

Hard Distil. \cite{radosavovic2017data} & \checkmark & & \checkmark & $S_{source}$ $U_{target}$ & $P_{source}$ $U_{target}$ & & & 51.5/75.7 & 48.7/76.4 & 68.4/67.2 & 20.6/92.0 & 44.3/77.8 \\
Naive Merging & \checkmark & \checkmark & & $S_{source}$ $U_{target}$ & $U_{source}$ $S_{target}$ & & & 37.0/82.8 & 31.9/84.3 & 55.7/75.2 & 7.7/93.6 & 28.1/83.7 \\

\hline

OMNIA Hard Distil.& \checkmark & \checkmark & \checkmark & $S_{source}$ $P_{target}$ & $P_{source}$ $S_{target}$ & & & 56.8/71.8 & 54.5/72.1 & 72.4/62.8 & 29.3/86.3 & 46.8/74.6 \\
OMNIA Discard Unsafe ROI & \checkmark & \checkmark & \checkmark & $S_{source}$ $P_{target}$ & $P_{source}$ $S_{target}$ & \checkmark & & 57.0/71.5 & 54.9/71.8 & 72.8/62.7 & 29.8/85.9 & 46.9/75.1 \\
OMNIA SoftSig & \checkmark & \checkmark & \checkmark & $S_{source}$ $P_{target}$ & $P_{source}$ $S_{target}$ & \checkmark & \checkmark & \textbf{57.4/71.0} & \textbf{55.2/71.6} & \textbf{74.6/62.0} & \textbf{30.4/85.6} & \textbf{48.3/74.3} \\

\hline

\end{tabular}
}

\caption{Complete ablation study. \textit{Source} dataset only has supervised ($S$) annotations for \textit{source} categories ($S_{source}$). \textit{Target} dataset is initially unsupervised for \textit{source} categories ($U_{source}$) while being supervised for the \textit{target} categories ($S_{target}$). Predictions ($P$) are used for replacing missing annotations. Only pictures with image-level labels are considered.}

\label{table:ablationfull}
\end{table*}

\begin{table*}
\centering
\resizebox{2.2\columnwidth}{!}{%
\small
\begin{tabular}{|c|| c c | c c c c c c c c c c c c|}
\hline
Method & \begin{tabular}{@{}c@{}}mean \\ mAP/MoLRP\end{tabular} & \begin{tabular}{@{}c@{}}mean without bag and tie \\ mAP/MoLRP\end{tabular} & \begin{tabular}{@{}c@{}}bag \\ AP/oLRP\end{tabular} & \begin{tabular}{@{}c@{}}belt\\ AP/oLRP\end{tabular} & \begin{tabular}{@{}c@{}}dress\\ AP/oLRP\end{tabular} & \begin{tabular}{@{}c@{}}footwear \\ AP/oLRP\end{tabular} & \begin{tabular}{@{}c@{}}headwear \\ AP/oLRP\end{tabular} & \begin{tabular}{@{}c@{}}outer \\ AP/oLRP\end{tabular} & \begin{tabular}{@{}c@{}}pants \\ AP/oLRP\end{tabular} & \begin{tabular}{@{}c@{}}tie \\ AP/oLRP\end{tabular} & \begin{tabular}{@{}c@{}}shorts \\ AP/oLRP\end{tabular} & \begin{tabular}{@{}c@{}}skirt \\ AP/oLRP\end{tabular} & \begin{tabular}{@{}c@{}}sunglasses \\ AP/oLRP\end{tabular} & \begin{tabular}{@{}c@{}}top \\ AP/oLRP\end{tabular}\\

\hline

Domain Adaptive (0) & 25.6/86.3 & 27.6/85.3 & 27.5/84.6 & 4.9/98.2 & 51.5/75.1 & 17.0/92.7 & 41.1/77.9 & 20.2/87.7 & 28.2/83.8 & 2.9/98.3 & 38.3/77.9 & 39.8/77.4 & 27.7/88.3 & 7.7/94.0 \\
Hard Distil. \cite{radosavovic2017data} & 31.3/84.7 & 27.4/86.4 & 42.4/79.9 & 6.2/98.2 & 50.8/76.4 & 17.6/92.8 & 39.6/84.0 & 19.0/88.0 & 28.6/83.9 & 59.5/73.2 & 35.9/78.7 & 39.3/79.7 & 28.5/88.2 & 8.2/93.8 \\
Naive Merging & 24.4/88.0 & 19.0/90.3 & 42.2/79.5 & 6.0/96.4 & 44.5/80.2 & 6.8/97.0 & 29.9/84.3 & 22.6/88.5 & 10.3/93.7 & 60.9/72.8 & 15.1/91.4 & 33.4/83.0 & 13.1/94.0 & 7.9/94.8 \\ 

\hline

OMNIA Hard Distil. & 36.9/81.1 & 33.7/82.1 & \textbf{43.5/79.3} & 6.9/97.1 & 60.5/\textbf{70.1} & 23.2/90.3 & \textbf{48.0/73.3} & 32.0/81.7 & 31.3/82.0 & \textbf{62.8/72.7} & \textbf{43.9/75.7} & 49.5/71.8 & 31.5/86.6 & 9.9/92.4 \\
OMNIA Discard Unsafe ROI & 36.1/81.5 & 32.9/82.7 & 43.3/78.6 & \textbf{10.2}/97.8 & 58.2/71.6 & 20.7/91.0 & 46.4/74.2 & 32.0/81.6 & 31.1/82.0 & 61.3/73.1 & 41.0/76.0 & 49.7/72.4 & 31.6/86.4 & 7.6/93.9 \\

OMNIA SoftSig & \textbf{37.2/80.8} & \textbf{34.4/81.7} & 43.2/79.1 & \textbf{10.2/95.7} & \textbf{61.0}/70.3 & \textbf{24.0/89.5} & 45.8/73.8 & \textbf{33.2/81.1} & \textbf{31.4/81.8} & 60.1/73.1 & 42.6/76.2 & \textbf{52.2/71.3} & \textbf{32.4/85.6} & \textbf{10.6/91.9} \\

\hline
\end{tabular}

}
\caption{Fashion detection results on \textit{OpenImages}. In addition to the pictures with image-level labels, we randomly sampled 10000 images from \textit{OpenImages} and assumed that they did not contain any garment. This assumption is often false but enables to unbias the selection procedure of the validation images.}
\label{table:modanetcocoopenrandom}
\end{table*}
\subsection{Experimental Setup}

Our code in Tensorflow \cite{tensorflow2015-whitepaper} is inspired from \cite{huang2017speed} \footnote{https://github.com/tensorflow/models/tree/master/research/object\_detection} and \cite{chen17implementation} \footnote{https://github.com/endernewton/tf-faster-rcnn}. As our approach does not rely on supplementary information regarding the datasets at hand, we adopt the training hyperparameters from Chen \textit{et al.} \cite{chen2018domain} \footnote{https://github.com/yuhuayc/da-faster-rcnn}. However, we use a Resnet101 \cite{he2016deep} as backbone, and a simpler domain classifier architecture without hidden layers because they did not bring significant improvements and were highly hyperparameters dependent.

A SGD optimizer with momentum 0.9 is used with a batch of 2 images, one from the first domain and the other from the second domain, without data augmentation. The batch size for RPN is 256. The box classifier batch size is 124: we sample 25\% of positive ROIs and 75\% of ROIs that are background or that match with an unsafe prediction. The weights are pretrained on ImageNet. The learning rate schedule depends on the task at hand, as explained below. In all our experiments $\lambda_{binary}$ is set to $1$. $threshold\_low$ is set to 0.2, $threshold\_high$ to 0.9 for all categories and we used class weights in training (rather than calibration) for better comparability across categories. Similarly to \cite{radosavovic2017data} for the object detection task and for simplicity, we choose not to use data distillation or model distillation. All the results are reported with a fixed seed, arbitrary set to 3, and compared with 2 complementary metrics: the \textit{mean Average Precision with IoU threshold at 0.5} (mAP@0.5) - the higher the better - and the \textit{Mean of optimal Localization Recall Precision} (MoLRP) \cite{LRP-ECCV18} \footnote{Through our experiments the MoLRP has shown to be a stable metric, but we will analyze the mAP for simplicity.} - the lower the better.

We conduct three main experiments: (\ref{chap:datasetmerging}) we show that our dataset merging procedure improves generalization results for \textbf{fashion detection for images in the wild}, (\ref{chap:newcategory}) we \textbf{add a new category in a dataset without relabelling} and (\ref{chap:domainshift}) we learn from \textbf{synthetic data with domain adaptation} using our soft distillation loss.

\subsection{Fashion for Images in the Wild}\label{chap:datasetmerging}

Detection of fashion items on images in the wild is challenging yet fundamental for many business applications such as the fashion trends understanding on social media or visual search.

\paragraph{Datasets \& Procedure}

In \textit{Modanet} \cite{zheng2018modanet}, 55,176 images were annotated for 13 categories (such as \textit{bag}, \textit{footwear}, \textit{dress}). However, these high resolution fashion images, with a single person posing, are limited in diversity. We leverage the currently biggest detection dataset with objects in context, \textit{COCO} \cite{lin2014microsoft}, with 80 objects (such as \textit{glasses}, \textit{firehydrant}) labeled on 117,281 training images. The scheduling from Tensorflow \cite{huang2017speed} is applied: the learning rate is reduced by 10 after 900K iterations and another 10 after 1.2M iterations. As \textit{COCO} is twice as big as \textit{Modanet}, the fashion images will be sampled 2 times more.

We evaluate our results on \textit{OpenImages} V4 validation dataset \cite{openimages}. Because of the large scale, the 125,436 images are only annotated for a category if this category has been detected by a image-level CNN classifier: that's why the ground truth annotations are not exhaustive. In the first experiment, we only consider the 13,431 pictures image-level labeled for at least one fashion garment. For example, only 1,412 pictures were verified at image-level for the category \textit{dresses} (1164 contain a dress, 248 do not), and contribute to the computation of the \textit{dress} AP.

\paragraph{Baseline \& Results}

Our experiments are summarized in Table \ref{table:ablationfull}. Our OMNIA detector is  better on all categories. It was expected for \textit{bag} and \textit{tie} as OMNIA benefits from additional annotations from \textit{COCO} (even though the labelling rules do not perfectly match) and we have a 6.3 points gain (+12.8\%) in average on all other categories.

Our first contribution is our \textbf{dataset creation procedure}. The Domain Adaptive \cite{chen2018domain} trained on \textit{Modanet} and adapted to \textit{COCO} does not generalize well to \textit{OpenImages} (see Figure \ref{fig:modanetvsomnia}). The Hard Distillation is trained similarly to \cite{radosavovic2017data} and use \textit{COCO} images as unlabelled data for bootstrapping. Finally, the Naive Merging's training datasets is the naive concatenation of initial datasets. Its low score confirms our initial intuition: the missing annotations of all fashion items in \textit{COCO} are detrimental. Our merging procedure enables to beat all three baselines by a large margin.

Our second contribution is the \textbf{handling of unsafe predictions}. The first component is the \textbf{masking}: in OMNIA Hard Distillation, unsafe predictions are considered as background (which leads to missing annotations) whereas these boxes are not sampled in OMNIA Discard Unsafe Predictions. The second component is the \textbf{binary cross entropy} that uses all available information by sampling even unsafe boxes and propagating safe gradients (Equation \ref{eq:binaryentropy}): it is the difference between the latter and OMNIA Softsig. Though this contribution is not as as impactful as the previous one, we still achieve a consistent gain over all of our experiments, which is more analyzed in next subsection.

\begin{table*}[b]
\centering

\resizebox{2.2\columnwidth}{!}{%
  \begin{tabular}{| c | c c || c | c | c  c  c c c c c ||c | c c|}
\hline
\multicolumn{3}{|c||}{} & \multicolumn{9}{c||}{Cityscapes val} & \multicolumn{3}{c|}{KITTI}\\
\hline
Method & SIM10k & Cityscapes & \begin{tabular}{@{}c@{}}mean \\ mAP/MoLRP\end{tabular} & \begin{tabular}{@{}c@{}}car \\ AP/oLRP\end{tabular} & \begin{tabular}{@{}c@{}}bike \\ AP/oLRP\end{tabular} & \begin{tabular}{@{}c@{}}bus \\ AP/oLRP\end{tabular} & \begin{tabular}{@{}c@{}}human \\ AP/oLRP\end{tabular} & \begin{tabular}{@{}c@{}}moto \\ AP/oLRP\end{tabular} & \begin{tabular}{@{}c@{}}rider \\ AP/oLRP\end{tabular} & \begin{tabular}{@{}c@{}}train \\ AP/oLRP\end{tabular} & \begin{tabular}{@{}c@{}}truck \\ AP/oLRP\end{tabular} & \begin{tabular}{@{}c@{}}car \\ AP/oLRP\end{tabular} & \begin{tabular}{@{}c@{}}human \\ AP/oLRP\end{tabular} & \begin{tabular}{@{}c@{}}truck \\ AP/oLRP\end{tabular} \\
\hline
\hline
Naive Merging & $S_{car}$ $U_{city}$ & $U_{car}$ $S_{city}$ & 30.0/88.2 & 4.9/97.8 & 36.9/87.0 & 37.9/80.6 & 35.0/87.1 & 30.1/89.3 & 47.7/81.8 & 23.9/93.9 & 23.8/88.7 & 28.3/87.6 & 44.4/83.1 & 17.8/90/7 \\
Oracle & - & $S_{car}$ $S_{city}$ & 40.9/83.6 & 57.5/70.8 & 34.6/88.7 & 53.8/74.7 & 35.0/87.4 & 30.3/89.9 & 42.9/87.5 & 37.3/87.1 & 36.1/83.0 & 72.1/65.9 & 49.3/82.7 & 23.7/89.3 \\
\hline
\hline
Domain Adaptive (1) \cite{chen2018domain} & $S_{car}$ $U_{city}$ & $U_{car}$ $U_{city}$ & -/- & 39.6/82.1 & -/- & -/- & -/- & -/- & -/- & -/- & -/- & 59.4/71.3 & -/- & -/-\\

\hline
OMNIA Hard Distil. & $S_{car}$ $P_{city}$ & $P_{car}$ $S_{city}$ & 40.6/83.6 & 42.8/80.6 & 40.2/85.5 & 50.1/75.6 & 37.7/85.8 & 31.8/89.6 & 48.0/80.2 & 38.4/88.7 & 36.0/83.1 & 67.8/68.3 & 53.2/80.9 & 25.6/89.4 \\
OMNIA Discard Unsafe ROI & $S_{car}$ $P_{city}$ & $P_{car}$ $S_{city}$ & 41.1/83.3 & 44.8/79.8 & 41.1/84.8 & 50.1/75.2 & 37.2/86.2 & \textbf{33.7/88.7} & 48.0/81.0 & 39.5/87.2 & 34.2/83.7 & 67.8/68.5 & \textbf{54.1/80.0} & 23.9/90.0 \\
OMNIA SoftSig (2) & $S_{car}$ $P_{city}$ & $P_{car}$ $S_{city}$ & 41.4/83.1 & 45.1/79.7 & 40.9/85.1 & 50.3/75.3 & 37.5/85.9 & 32.3/89.8 & 48.3/80.4 & \textbf{40.3/86.3} & 36.6/82.6 & 68.7/67.8 & 53.7/80.0 & 25.8/88.7 \\
\hline
OMNIA SoftSig Iterative & $S_{car}$ $P_{city}^{iter}$ & $P_{car}^{iter}$ $S_{city}$ & \textbf{42.2}/\textbf{82.7} & \textbf{45.3/79.6} & \textbf{41.9/84.9} & \textbf{50.4/75.2} & \textbf{37.8/85.5} & 33.0/87.9 & \textbf{51.1/79.7} & 42.1/86.5 & \textbf{36.9/82.5}  & \textbf{69.9/67.3} & 53.8/79.9 & \textbf{25.9/88.7}  \\
\hline
\hline
Oracle SoftSig Iterative & $S_{car}$ $P_{city}^{iter}$ & $S_{car}$ $S_{city}$ & 44.6/80.8 & 60.5/67.2 & 41.1/84.8 & 53.2/74.5 & 38.5/85.4 & 35.5/87.6 & 49.6/79.7 & 42.4/85.6 & 36.1/82.1 & 75.5/63.2 & 53.7/79.8 & 26.5/87.4 \\
\hline

\end{tabular}
}

\caption{Results on \textit{Cityscapes} and \textit{KITTI}. \textit{SIM10k} only provides supervised ($S$) annotations for \textit{cars} ($S_{car}$). \textit{Cityscapes} is initially unsupervised ($U$) for \textit{cars} ($U_{car}$) while being supervised for the seven other \textit{city} categories ($S_{city}$). Predictions ($P$) from detector (1) compensate the \textit{cars} missing annotations on \textit{Cityscapes} ($P_{car}$), and can be improved in an iterative process ($P^{iter}$) from detector (2).}

\label{table:sim10kcitymergingcityandkitti}
\end{table*}

\paragraph{Unbiased but approximate results on \textit{OpenImages}}

There is a bias in the selection method, as these images are annotated only if an image-level classifier predicted that they contained a fashion item and are therefore fashion-oriented. In order to better understand the benefit of our procedure, we sampled 10,000 images randomly from \textit{OpenImages} and suppose that they did not contain any garment. This assumption is not fully correct, but reasonable as there are usually few garments classes when images are chosen randomly. They enable to reduce the bias in the selection procedure. For each garment, the images of the final test set (to be released for further comparison) are either image-level labeled or sampled randomly.
The results are reported in Table \ref{table:modanetcocoopenrandom}. The performance gap is even larger (+6.8 points, +24.6\%) because these random images are not fashion oriented: some of them contain \textit{fire hydrants} and \textit{bottle} on which the Domain Adaptive often makes highly confident yet wrong predictions \cite{li2018mathcal} (see Figure \ref{fig:modanetvsomnia}).

\paragraph{Future work}
Our methodology could be used for direct training from \textit{OpenImages} while handling missing annotations. Moreover, a better selection metric to separate safe and unsafe predictions would help: this metric should be independent from the initial models and should have \textit{orthogonal} failure modes \cite{rosenberg2005semi}. Lastly, improved losses could handle uncertainty without binarizing the prediction score.



\paragraph{Dataset Size Analysis}

\begin{figure}[]
\centering
\includegraphics[width=0.5\textwidth]{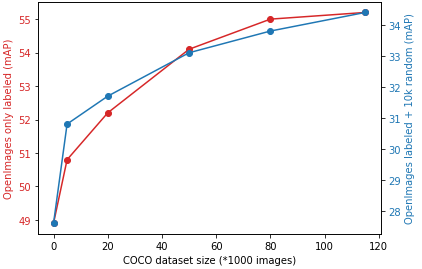}
\caption{Influence of the size of the subset of \textit{COCO} over the final mAP@0.5 for the two validation datasets (1) with only human labeled images (2) with additional random non-fashion images.}
\label{fig:sim10kcity_threshold}
\end{figure}

To better understand the importance of the number of images from \textit{COCO}, we apply our methodology using only subsets of different sizes of the full training dataset (see Figure \ref{fig:sim10kcity_threshold}). As expected, increasing the number of images labeled from \textit{COCO} improves the final performance. This gain is not linear but logarithmic: a small subset of \textit{COCO} is already useful, even more for the validation dataset with random images from \textit{OpenImages}. In conclusion, we could merge two relatively small datasets and they would benefit from each other as well.



\subsection{Adding a New Category without Relabelling}\label{chap:newcategory}

Adding a new entity is usually expensive as a complete labelling throughout the dataset is required. We investigate the possibility to leverage another dataset where this entity is already labeled: it is a special case of dataset merging where the first dataset has only one category ($|\mathcal{C}_{a}|$ = 1).

\paragraph{Datasets \& Procedure}
Our goal is to detect \textit{cars} in real-world images, having only instances from a synthetic dataset with a clear domain shift in terms of image quality and resolution. \textit{SIM10k} \cite{johnson2016driving} provides 10,000 training images with 58,701 \textit{cars} automatically annotated from the video game \textit{Grand Theft Auto}. The 2,975 images from \textit{Cityscapes} \cite{cordts2016cityscapes} are also given, and are labeled for seven categories (\textit{bicycle}, \textit{bus}, \textit{human}, \textit{motorcycle}, \textit{rider}, \textit{train}, \textit{truck}): the \textit{car} annotations from these images are filtered. We apply the dataset merging procedure on \textit{Cityscapes} and \textit{SIM10K}, with a learning rate of 0.002 for 15k iterations which is then reduced to 0.0002 for another 15k iterations.

\paragraph{Results on \textit{Cityscapes}}

\begin{figure}
\begin{center}
   \includegraphics[width=0.99\linewidth]{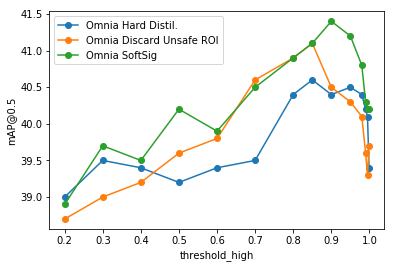}

   \end{center}
\caption{Threshold impact on the final map during the merging of \textit{Sim10k} and \textit{Cityscapes}.}
\label{fig:threshold}
\end{figure}

Table \ref{table:sim10kcitymergingcityandkitti} summarizes the results of the different methods on \textit{Cityscapes} validation. The Naive Merging have really poor results, as \textit{cars} in \textit{Cityscapes} are considered as background.The Domain Adaptive trained on \textit{SIM10k} reaches 39.6: predictions of this model on \textit{Cityscapes} are used as targets in the final training. Considering the unsafe predictions as background in OMNIA Hard Distillation attains 40.6 mAP. We reach 41.1 mAP when we simply discard these unsafe regions by not sampling unsafe ROIs. The OMNIA SoftSig score of 41.4 mAP is slightly higher thanks to using all available information. In particular, the category \textit{trucks} largely benefits from the SoftSig. As \textit{trucks} and \textit{cars} are similar in appearance, the unsafe predictions of \textit{cars} are often regions where the model could predict \textit{trucks}. Explicitly learning that these regions are not \textit{trucks} improves its AP from 34.2 to 36.6. The SoftSig is better independently of the threshold (see Figure \ref{fig:threshold}). The improvement generalizes well across the seven other categories, as more labeled examples yield better generalization: as noticed in \cite{radosavovic2017data}, we \textit{learn new knowledge from the extra unlabeled data}. The best score of 42.2 mAP is obtained for OMNIA SoftSig Iterative, where the procedure is repeated iteratively using the OMNIA Softsig's predictions as targets. Results are there close to the results obtained by the Oracle, where \textit{cars} are not removed from \textit{Cityscapes}.

\paragraph{Results on \textit{KITTI}}

We compare the generalization ability of our detector on a new domain. \textit{KITTI} \cite{geiger2013vision} contains 7,481 real-world images, but with a different data collecting method than \textit{Cityscapes}: the performances in \textit{Cityscapes} do not perfectly match those in \textit{KITTI}. In table \ref{table:sim10kcitymergingcityandkitti} we see that our OMNIA has 69.9 of AP for \textit{car}, which is close to the Oracle AP of 72.1. The performance gap for \textit{car} in the last experience (45.3 vs 57.3) was mainly due to a domain specialization. Learning to detect \textit{car} from 2 domains, \textit{SIM10k} and \textit{Cityscapes}, induces a better domain invariance. We also notice that our SoftSig detector has a better \textit{cars} AP than the Hard Distillation and a better \textit{truck} AP than when we do not sample unsafe regions.



\subsection{Learning from Synthetic Data with Domain Adaptation}\label{chap:domainshift}

Domain adaptation is equivalent to the problem of dataset merging when one dataset has no labeled classes ($|\mathcal{C}_{b}|$ = 0). This case is used for comparison with previous works \cite{chen2018domain, shan2018pixel}. Therefore  in this experiment the unsupervised domain adaptation protocol is adopted. The training data consists of two parts: (1) the source dataset (\textit{SIM10k}) where images and their associated instances annotations are provided and (2) the target dataset (\textit{Cityscapes}) with images without supervision. Our procedure aims at detecting robustly on the \textit{Cityscapes} validation by using unlabeled images from \textit{Cityscapes} training. Note that in this experiment we only consider \textit{cars} and discard the seven other \textit{city} categories. Our detectors are trained with a learning rate of 0.002 for 10,000 iterations which is reduced to 0.0002 for another 10,000 iterations.

\begin{table}
\centering

\resizebox{1.\columnwidth}{!}{%

\begin{tabular}{|c | c | c c || c c |}
\hline
Method & Backbone & SIM10k & City & \begin{tabular}{@{}c@{}} City car \\ AP/oLRP\end{tabular} & \begin{tabular}{@{}c@{}} KITTI car \\ AP/oLRP\end{tabular}\\
\hline
\hline
Baseline & R101 & $S_{car}$ & - & 34.4/84.0 & 55.0/78.5 \\
Domain Adaptive \cite{chen2018domain} & VGG & $S_{car}$ & $U_{car}$ & 39.0/- & -/- \\
Pixels DA \cite{shan2018pixel} (SOTA) & VGG & $S_{car}$ & $U_{car}$ & 39.6/- & -/- \\
Oracle & R101 & - & $S_{car}$ & 59.6/69.7 & 71.1/66.5 \\
\hline
\hline
Our Domain Adaptive (1) \cite{chen2018domain} & R101 & $S_{car}$ & $U_{car}$ & 39.6/82.1 & 59.4/71.3\\
\hline

OMNIA Hard Distil. & R101 & $S_{car}$ & $P_{car}$ & 43.1/80.9 & 64.2/70.6 \\
OMNIA Softsig & R101 & $S_{car}$ & $P_{car}$ & \textbf{44.9}/\textbf{80.0} & \textbf{65.4/70.1} \\
\hline
\end{tabular}
}
\caption{Domain adaptation results on \textit{Cityscapes} and \textit{KITTI}. \textit{SIM10k} is the source dataset. \textit{car} predictions $P_{car}$ from detector (1) on \textit{Cityscapes} are used for self-training. See Table \ref{table:sim10kcitymergingcityandkitti} for more notation conventions.}
\label{table:sim10kcityweakempty}

\end{table}


Table \ref{table:sim10kcityweakempty} presents our results. Our baseline is trained purely on the source \textit{SIM10k} dataset with a mAP of 35.4. The Oracle trained with labeled \textit{Cityscapes} reaches 59.6. We reproduced the work from \cite{chen2018domain} (39.6 vs 39.0 with a VGG but better hyperparameters). A \textit{hard} distillation procedure reaches 43.1. Our soft distillation achieves 44.9 which is a new SOTA with a gain of +5.3 points. As there is only one category to detect, ROIs that match with unsafe boxes do not provide any additional information: they are not sampled. Self-supervision with soft distillation can improve domain adaptation. Our procedure also improves \textit{car} AP on \textit{KITTI}. Interestingly, our results for \textit{car} are lower than when we have full annotations in \textit{Cityscapes} for the seven other categories (65.4 vs 68.7, see Table \ref{table:sim10kcitymergingcityandkitti}): as previously stated, being able to detect objects such as \textit{trucks} enable to reduce the number of false positive predictions.

\section{Conclusion}

In this paper, we presented our dataset merging procedure and our soft classification loss that enable to train robust detectors from small and specific datasets. The final detector is able to benefit from distinct datasets and predicts all categories with no additional labelling. By enabling the training on many more categories, we improve our definition of what background is. 

We think that learning detectors from numerous weakly supervised datasets is a critical step towards open set object detection, and hopefully towards human-level reliability.

\paragraph{Acknowledgements}
We would like to thank Arthur Douillard and Jason de Rancourt for their valuable and constructive suggestions during the writing of this paper. More generally, this work would not have been possible without the friendly support from all our co-workers at Heuritech.


{\small
\bibliographystyle{ieee}
\bibliography{biblio}
}

\end{document}